\documentclass[runningheads]{llncs}
\usepackage{graphicx}

\usepackage{tikz}
\usepackage{comment}
\usepackage{amsmath,amssymb} 
\usepackage{color}
\usepackage{orcidlink}

\usepackage[accsupp]{axessibility}  

\usepackage{graphicx}
\usepackage{booktabs}
\usepackage{multicol,multirow}
\usepackage{bbm}
\usepackage{soul}
\usepackage{enumitem}

\setlist{nosep}
\usepackage[normalem]{ulem}
\usepackage{caption}
\usepackage{subcaption}
\usepackage{url}
 \usepackage{floatrow}
\begin{document}
\pagestyle{headings}
\mainmatter
\def\ECCVSubNumber{100}  

\title{Towards Open-vocabulary Scene Graph Generation with Prompt-based Finetuning } 

\titlerunning{Towards Open-vocabulary SGG with Prompt-based Finetuning}
%
\author{Tao He\inst{1} \and
Lianli Gao\inst{2} \and
Jingkuan Song\inst{2}\and Yuan-Fang Li \inst{1}$^\dag$} 
\authorrunning{ He et al.}
%
\institute{Faculty of Information Technology, Monash University,  \and
Center for Future Media, University of Electronic Science and Technology of China 
  \\
\email{\{tao.he, yuanfang.li\}@monash.edu, lianli.gao@uestc.edu.cn, jingkuan.song@gmail.com}}
{\let\thefootnote\relax\footnotetext{$\dag$ Corresponding author.}}

\maketitle

\begin{abstract}
Scene graph generation (SGG) is a fundamental task aimed at detecting visual relations between objects in an image. The prevailing SGG methods require all object classes to be given in the training set. Such a closed setting limits the practical application of SGG. In this paper, we introduce \emph{open-vocabulary} scene graph generation, a novel, realistic and challenging setting,  in which a model is trained on a set  of base object classes but is required to infer relations for unseen target object classes. To this end, we propose a two-step method which firstly pre-trains on large amounts of coarse-grained region-caption data  and then leverages two prompt-based techniques to finetune the pre-trained model without updating its parameters. Moreover, our method is able to support inference over completely unseen object classes, which existing methods are incapable of handling. On extensive experiments on three benchmark datasets, Visual Genome, GQA and Open-Image, our method significantly outperforms recent, strong SGG methods on the setting of Ov-SGG, as well as on the conventional closed SGG.
 
\end{abstract}

 \section{Introduction}
 Scene Graph Generation (SGG)~\cite{tang2019learning,zellers2018neural,suhail2021energy,tang2020unbiased,zareian2020bridging} aims at generating visual relation triples in a given image and is one of the fundamental tasks in computer vision. It has wide applications in a suite of high-level image understanding tasks, including visual captioning~\cite{yao2018exploring,ye2021linguistic,yang2019auto}, visual question answering~\cite{teney2017graph,antol2015vqa}, and 3D scene understanding~\cite{armeni20193d,zhang2021exploiting,he2021exploiting}. 
 
 It was not until recently that  {the long-tail   distribution in SGG datasets}   was identified~\cite{zellers2018neural}. Following this discovery, a number of works~\cite{tang2019learning,chen2019knowledge,he2020learning,li2021bipartite,wang2020tackling,Lin_2020_CVPR} endeavoured to reduce the impact of the biases in data by exploiting debiasing techniques~\cite{tang2020unbiased,li2021bipartite,Knyazev_2021_ICCV}. Although remarkable improvements in the performance of unbiased SGG have been made, these state-of-the-art methods are limited to predicting relationships between pre-defined object classes only. In real-world scenarios, however, it is highly likely that an SGG model encounters objects of \emph{unseen} categories that do not appear in the training set. In this more realistic and practical setting, 
 \begin{figure}[!ht]
 	\floatbox[{\capbeside\thisfloatsetup{capbesideposition={right,top},capbesidewidth=0.35\linewidth}}]{figure}[\FBwidth]
 	{\caption{An illustration of the conventional closed SGG $vs.$ Ov-SGG. For the unseen target object \texttt{boot}, closed SGG methods such as EBM~\cite{suhail2021energy} and Motifs~\cite{zellers2018neural}, cannot predict any relation regarding \texttt{boot} whilst our Ov-SGG method is able to. }  \label{fig:1}}
 	{\includegraphics[width=1\linewidth]{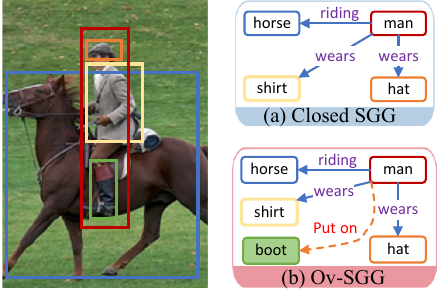}}
 \end{figure}
 the performance of these conventional SGG models~\cite{zellers2018neural,tang2020unbiased,Knyazev_2021_ICCV} degrades, and dramatically so when inferring over completely unseen object classes, {as can be seen in Tab.~\ref{tab:1} in Sec.~\ref{sec:exp})}. 
 This motivates us to ask the question,  \textit{can we predict visual relationships for unseen objects?} 
 Formally, we call this problem setting \textbf{Open-vocabulary Scene Graph Generation} (Ov-SGG).

 
 In Ov-SGG, the model is trained on objects belonging to a  set of seen (i.e.\ base)  object categories {$\mathcal{O}^b$} and then predicts relationships on unseen (i.e.\ target) object categories {$\mathcal{O}^t$}, both of which are subsets of the open-vocabulary object class set $\mathcal{O}=\mathcal{O}^b\cup\mathcal{O}^t$. This is distinct from, and more challenging than, most existing zero-shot scene graph generation (Zs-SGG) settings~\cite{lu2016visual} or weakly supervised scene graph generation (Ws-SGG) settings~\cite{zareian2020weakly}. 
 More specifically, Zs-SGG~\cite{lu2016visual} is dedicated to predicting the relationship of objects whose combinations do not emerge in the training set, where the objects are all from seen categories. {In contrast, in Ov-SGG, at inference time, not only object combinations may be novel, object categories themselves may not have been seen by the model during training.} Zhong \textit{et al.} \cite{zhong2021learning} extends weakly supervised scene graph generation~\cite{zareian2020weakly} to learning scene graphs from image-caption pairs and  demonstrates their model's capability of {open-set} SGG. In their open-set configuration, the model is trained with a large object set ($4,273$ classes) and a predicate set ($677$ categories), and the object categories for testing are included in the training object set. 
 In other words, both settings assume $\mathcal{O}^t \subseteq \mathcal{O}^b$ should be seen beforehand. In contrast, in our open-vocabulary setting, the model is presented with some unseen object classes at inference time, i.e., $\mathcal{O}^t\setminus\mathcal{O}^b\neq \varnothing$, and in the extreme case, completely unseen object classes, i.e., $\mathcal{O}^t\cap\mathcal{O}^b=\varnothing$. 
 Learning scene graphs for unseen object categories has so far remained unexplored. {Fig.~\ref{fig:1}  illustrates a comparison between Ov-SGG and its closed-world counterpart.} Moreover, we propose a more realistic and more challenging setting, in which the test set contains novel relation predicates not seen during training. We name this task as \emph{general open-vocabulary} SGG, or gOv-SGG. For instance, in Fig.~\ref{fig:1}, the novel relation ``put on'' makes this task gOv-SGG. 
 
 The main challenge for Ov-SGG is the knowledge gap between the base and target object categories, i.e., how to leverage the learned visual patterns from the limited base categories for the target categories. To bridge this gap, we propose a two-step method of \emph{\textbf{visual-relation pre-training}} and \emph{\textbf{prompt-based finetuning}}.
 
 Firstly, we capitalize on a large number of   visual-textual  pairs to pre-train a cross-modal model to align the visual concepts with their corresponding unbounded textual descriptions. Different from  visual-language models~\cite{zhang2021vinvl,su2019vl,radford2021learning} that {are pre-trained on whole images and their captions}, we leverage the dense captions of Visual Genome that focus on detailed regional semantics, as we believe that the dense captions can provide more localised relational information. 
 Secondly, we design two prompt-based learning strategies, hard prompt and soft visual-relation prompt. The pre-trained model makes predictions by \emph{filling in the blank} in the designed prompt. 
 
 Finetuning has been widely employed to further reduce the knowledge gap between the pre-trained model and downstream tasks \cite{su2019vl,zhang2020contrastive,li2020oscar} . However, the standard finetuning practice does not lead to promising results on Ov-SGG, as the newly introduced task-specific prediction heads cannot well handle unseen scenarios, as observed recently \cite{liu2021pre}.
 Prompt-based learning \cite{chen2021adaprompt,li2021prefix,davison2019commonsense} has enjoyed remarkable success in a variety of downstream tasks in natural language processing, including relation extraction~\cite{chen2021adaprompt}, commonsense knowledge mining~\cite{davison2019commonsense} and text generation~\cite{li2021prefix}. 
 This is achieved by learning a small amount of parameters for prompt generation without the need to update parameters of the large underlying pre-trained model. As a result, compared to standard finetuning, prompt-based learning suffers less from task interference and thus enjoys better zero-shot learning ability . 
 
 Our contributions can be summarized as follows.
 \begin{itemize}[nosep]
 	\item We propose the new, practical and challenging tasks of \emph{open-vocabulary scene graph generation} (\textbf{Ov-SGG}), together with a more challenging setting, \emph{general} open-vocabulary SGG (\textbf{gOv-SGG}). We believe that Ov-SGG and gOv-SGG represent a firm step towards the real application of SGG.
 	\item We propose a two-step method that firstly \emph{\textbf{pre-trains a visual-relation model}} on large amounts of available region-caption pairs aiming for visual-relation alignment with open-vocabulary textual descriptions, and secondly finetunes the pre-trained model by two \emph{\textbf{prompt-based finetuning strategies}}: hard prompt and soft visual-relation prompt. To our best knowledge, this is the first investigation of prompt-based finetuning for SGG. 
 	\item Extensive experiments on three standard SGG datasets, VG~\cite{krishna2017visual}, GQA~\cite{hudson2019gqa} and Open-Image~\cite{kuznetsova2020open}, demonstrate our model's significant superiority over state-of-the-art SGG methods on the task of Ov-SGG. Moreover, our method is the only
 	one capable of handling the more challenging zero-shot object SGG. Finally, our model also consistently surpasses all compared baselines on the standard closed SGG task.
 \end{itemize}

 \section{Related Work}
 \noindent \textbf{Scene graph generation (SGG)}~\cite{chang2021scene} aims to detect and localise visual relation between objects.   Primary works~\cite{schuster2015generating,johnson2015image,he2020learning} mainly view scene graphs as auxiliary information to improve the quality of image retrieval. Later on, a couple of following arts \cite{yao2018exploring,yang2019auto,teney2017graph} demonstrate that scene graphs can be applied to various visual tasks and  significantly improve their performance, particularly because scene graphs can provide those models with structured visual representations and facilitate image understanding. With the standard SGG benchmark dataset Visual Genome~\cite{krishna2017visual} coming  forth for the public, a number of researchers~\cite{xu2017scene,zellers2018neural,zhang2017visual,he2021semantic,tang2019learning,chen2019knowledge,suhail2021energy} started put their efforts to SGG. Xu \textit{et al.} \cite{xu2017scene} leveraged an iterative message passing technique derived from graph convolution network~\cite{kipf2016semi} to refine   object features and improve the quality of generated scene graphs.   Yang \textit{et al.} \cite{yang2019auto} developed an auto-encoder network  by incorporating image-caption into SGG. Zeller \textit{et al.} 
 \cite{zellers2018neural} first pointed out the bias of relation distribution in the VG dataset and revealed that even using the statistic  frequency can obtain the comparable   SOTA performance at that time. Thus, a suit of subsequent works \cite{tang2020unbiased,tang2019learning,chen2019knowledge,Lin_2020_CVPR,suhail2021energy} started to work on tackling the tricky  bias problem in SGG. 
 However, many researchers ignore  a common issue that is  the SGG models' generalizability.  Although  some works~\cite{ye2021linguistic,baldassarre2020explanation,peyre2017weakly} proposed  the task of weakly supervised scene graph generation, they typically focus on generating scene graphs without localisaed clues, e.g., from image-caption pairs~\cite{ye2021linguistic} or VCR~\cite{zellers2019vcr}.
 
 \noindent \textbf{Pretrained visual-language (VL) models} have been widely applied for tremendous visual-language tasks \cite{su2019vl,zhang2020contrastive,wang2020vd,radford2021learning}. Without of loss generality, we could divide those works into two stages: (1)  a cross-modal model is first pretrained by natural supervision, e.g., image-caption pairs,  to align   visual features with their corresponding textual semantics in the common space by a contrastive learning \cite{radford2021learning,zhang2020contrastive} or masked token loss \cite{su2019vl,zhang2021vinvl}, and (2) developing a suite of visual-language fusion mechanisms, e.g., concatenation of both features~\cite{deng2021transvg}, and then finetuning  the pretrained  VL model for downstream  VL tasks, such as image caption~\cite{su2019vl} and visual dialog \cite{wang2020vd}. Recently, CLIP \cite{radford2021learning} used massive image-text data to train two encoders for images and texts solely by a simple contrastive loss and showed promising results on a wide-range of image classification tasks, especially in the zero-shot scenario.

 \noindent \textbf{Prompt-based learning} \cite{liu2021pre} has gained extensive attention in natural language processing due to successful applications of large-scale pretrained language models such as Bert \cite{devlin2018bert} and GPT-3 \cite{brown2020language}. Early studies~\cite{radford2018improving,peters2018deep} typically focused on  finetuning the pretrained language models to adapt the model to different downstream tasks by training new parameters via  task-specific objective functions.  With the advent of prompt-based  learning \cite{trinh2018simple}, following works \cite{chen2021adaprompt,li2021prefix,davison2019commonsense}  turn to prompt engineering  by designing appropriate prompts for the pretrained language model without adding task-specific training or modifying the language parameters.   Chen \textit{et al.}~\cite{chen2021adaprompt} develops an adaptive prompt finetuning strategy for entity relation extraction. Li \textit{et al.} \cite{li2021prefix} proposes a lightweight prefix-tuning strategy which enables the mode to fix the parameters of the pretrained language model during training while  learns   continuous task-specific vectors for text generation.
 

 \section{Problem definition}\label{sec:pd}
 Scene graph generation (SGG) aims to detect and localize the relationship between objects in an image.  
 \emph{Open-vocabulary} scene graph generation (\textbf{Ov-SGG}) is a more challenging problem that works in a more generic scenario, i.e., trained on a  set of base object but predicting the relationship for target objects. 
 Formally, let us consider a set of  {base} object categories  $\mathcal{O}^b$ where $c_k$ denotes a semantic label of an object. To model the realistic open-world scenario, we assume that there also exists another set of  {target} object classes $\mathcal{O}^t$, where $\mathcal{O}^t \setminus \mathcal{O}^b \neq \emptyset$, i.e., the target set $\mathcal{O}^t$ contains \textit{novel classes} not found in the base set $\mathcal{O}^b$. Moreover, all entities in relation triples in the training set are labeled only by the base object classes in $\mathcal{O}^b$, i.e., $\mathcal{D}^t=\{(x_i,\boldsymbol{y}_i)\}_{i=1}^M$, where $M$ is the number of training images; $x_i$ denotes the $i$-th image and $\boldsymbol{y}_i$ is the corresponding scene graph that comprises the set of $n_i$ annotated relation triples, i.e., $\boldsymbol{y}_i = \{(s_j,r_j,o_j)\}_{j=1}^{n_i}$ where  $s_j,o_j \in \mathcal{O}^b$ are the subject and object respectively; $r_j \in \mathcal{R}$ is the relation predicate; and $\mathcal{Y}$ is the set of all relation predicate words. 
 
 The goal of Ov-SGG is to train a model $\mathcal{M}$ on $\mathcal{D}^b$ so that during the inference stage, $\mathcal{M}$ can predict not only relations of $\mathcal{O}^b \times \mathcal{O}^b $ pairs, but also of {$\mathcal{O} \times \mathcal{O} $},  where $\mathcal{O}=\mathcal{O}^b \cup \mathcal{O}^t$ and  $\times$ is the Cartesian product operation. At the same time, 
 a derivative task, zero-shot object SGG (\textbf{ZsO-SGG}) requires a model $\mathcal{M}$ trained on $\mathcal{O}^b$ to predict over $\mathcal{O}^t\times\mathcal{O}^t$, where additionally, $\mathcal{O}^t$ contains novel classes only, i.e., $\mathcal{O}^t\cap\mathcal{O}^b=\emptyset$. We also define a more challenging setting, general Ov-SGG (\textbf{gOv-SGG}), in which the predication set $\mathcal{R}$ contains novel predicates in the testing stage that are not seen in training. I.e., $\mathcal{R} = \mathcal{R}^b\cup\mathcal{R}^t$ and $\mathcal{R}^t\setminus\mathcal{R}^b\neq\emptyset$.

 \section{Method}
 \begin{figure*}[htb]
 	\centering
 	\includegraphics[width=1\linewidth]{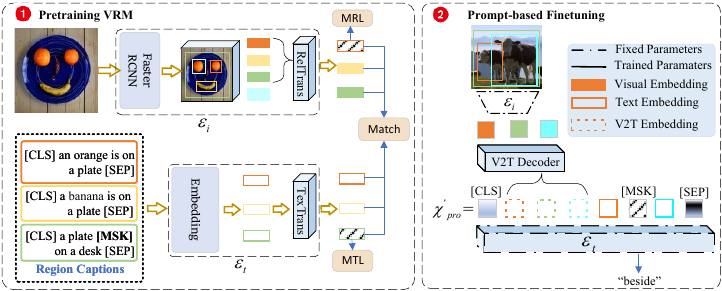}
 	\caption{ 
 		The overview of our framework for Ov-SGG. \textbf{Left}:   the architecture of pretraining VRM, which consists of two encoders for regions and dense captions in VG  trained by three losses: masked region loss (MRL), match loss and masked token loss (MTL). \textbf{Right}:  our proposed soft visual-relation prompt for finetuning, which leverages the off-the-shell VRM and optimizes a visual to textual (V2T) decoder network to produce soft prompts.
 	}\label{fig:frameword}
 \end{figure*}
 
 Fig.~\ref{fig:frameword} shows the overview of our Ov-SGG framework, which is based on a pre-trained visual-relation model (VRM) by tremendous visual-textual corpus, and finetuned  on the base objects  in a prompt-based learning manner. More specifically, we first train two transformer encoders for image and text, as in \cite{radford2021learning}, to align the visual concepts with their open-vocabulary  {relational description}. With the pretrained  VRM, we devise two prompt-based finetuning strategies to reduce the knowledge gap between VRM and SGG   by   a fill-in-the-blank paradigm.

 \subsection{Pretrained context-aware visual-relation model}
 An intuitive idea to pretrain a visual-relation model is to use the relation triples  subject-predicate-object (SPO) comprised of $\mathcal{O}^b$ to train such a visual-relation  space. However, this practice could result in heavily overfitting issue due to the small number of $\left|\mathcal{O}^b\right|$. Thus, we consider to use unbounded vocabulary relation corpus to train the model. Additionally,  the majority of VL models  \cite{radford2021learning,huang2020pixel} totally attempt to align visual concepts with their global textual semantic from   image-caption pairs in a fixed scheme, but the identical object in an image could have different relation in an SG, which depends on the other object in a pair. Hence, the pre-trained model should be able to map a visual component into various relation concepts according to its regional context. This ability is not available for the previous pretrained models~\cite{radford2021learning,li2020oscar,huang2020pixel}. 
 Toward this goal, we propose to train a regional context-aware visual-relation model on the dense-caption of VG by two Transformer-based encoders for images and texts (left of Fig.~\ref{fig:frameword}).

 \noindent \textbf{Training data.} We collect the dense-caption in the {training} set of Visual Genome (VG) as our pretrained data with  $\sim\! 2.6$ million region-caption pairs with bounding box information. More specifically, each image on average consists of $\sim\!\!50$ various dense region descriptions. Note that the dense caption is different from the image caption, e.g., MSCOCO~\cite{lin2014microsoft} with a general   global structured description, while  the dense caption generally depicts an object's relation with another object within a region.  In VG, the vast majority of region captions describe relationships. For more dataset statistic details, please refer to \cite{krishna2017visual}.  

 \noindent {\textbf{Image Encoder}}. 
 The image encoder ($\mathcal{E}_i$ in Fig.~\ref{fig:frameword}) consists of two modules: a  region proposal feature extractor (e.g., Faster-RCNN~\cite{ren2015faster}) and a  relation Transformer embedding module. The Transformer network takes the region proposals ${\boldsymbol{r}}$ as input visual tokens. 
 Since our model attempts to encode the relation-specific regional context, we do not feed all regions into the encoder like in previous works~\cite{radford2021learning,su2019vl,zhang2021vinvl}. Instead, we propose a union region based sampling strategy that 
 samples feasible regions as the regional context. 
 Concretely, we first randomly sample two anchor regions as  the {top left} ${r_t}$ and {bottom right} ${r_b}$ region 
 and then select the other regions ${[r_1...r_m]}$ that overlap with the union region $\operatorname{Union}(r_t,r_b)$ as the context. Here, we set an IoU threshold to select those regions. Briefly, we formulate the image encoding processing as follows: 
 \begin{equation}
 	\mathbf{h} = \operatorname{RelTrans}(\boldsymbol{W}\!_1  [r_t,r_1, \ldots, r_m,r_b]+\boldsymbol{l})
 	\label{eq:trn}
 \end{equation}
 where $\mathbf{h}=\mathrm{[h_t,h_1...h_m,h_b]}$ are the embeddings for each visual token; $\operatorname{RelTrans}(\cdot)$ is the relation transformer module; $\boldsymbol{W}\!_1$ is a learnable projection matrix; and $\boldsymbol{l}$ is the  {positional embedding} \cite{radford2021learning} for each token. 
 
 \noindent {\textbf{Text Encoder.}} 
 The text encoder ($\mathcal{E}_t$ in Fig.~\ref{fig:frameword}) is a parallel Transformer branch to the image encoder, which takes the corresponding region description as the input and produces their embeddings by:
 \begin{equation}
 	\boldsymbol{e}_i\! =\! \operatorname{TexTrans}(\boldsymbol{W}\!_2  \texttt{[CLS}, w_1, w_2, \ldots, w_k,\texttt{EOF]}\! +\!\boldsymbol{l}^\prime)
 	\label{eq:text_embed}
 \end{equation}
 where $\boldsymbol{c}{_i}=\left[ w_1, w_2, \ldots, w_k \right]$ denotes   $k$ words  in the dense-caption of region $r_i$; $\boldsymbol{l}^\prime$ is the position embeddings for each token; and $\texttt{[CLS]}$ and $\texttt{[EOF]}$ are  learnable special tokens, denoting the first  and last word, respectively. Note that we choose the embedding of $\texttt{[EOF]}$ as the   embedding of $\boldsymbol{c}{_i}$.

 \noindent \textbf{Pre-trained Loss Function.}  The training loss is designed from two main perspectives: image-text matching loss~\cite{zareian2021open} and masked token loss~\cite{li2020oscar}. For the former, we   adopt the cosine contrastive loss $\mathcal{L}_\text{c}$    \cite{radford2021learning} to force the visual region embedding to match the embedding of its corresponding  dense-caption.
 For the latter, regarding to $\operatorname{RelTrans}$, we mask any region in $\mathbf{h}$, i.e., replacing it by the special token $\texttt{[mask]}$, with $15\%$ probabilities 
 and let $\mathrm{h}_\mathrm{[mask]}$ be close to the embedding of its ground-truth caption by the contrastive loss, denoted as masked region loss $\mathcal{L}_\mathrm{MRL}$. As for  the text encoder, we follow the standard masked language models~\cite{zareian2021open,wang2020vd} to train $\mathrm{TexTrans}$ by a cross-entropy loss, which we denote by $\mathcal{L}_\mathrm{MTL}$. The total pre-trained loss is defined as:
 \begin{equation}
 	\mathcal{L}_{pre} = \mathcal{L}_\text{MRL}+\mathcal{L}_\text{MTL}+\mathcal{L}_\text{c}
 	\label{eq:pre_loss}
 \end{equation}
 %
 %
 \subsection{Prompt-based Finetuning for Ov-SGG} \label{sec:finetune}
 
 In this section, we introduce our proposed prompted-based finetuning method for Ov-SGG, which exploits the rich visual relationship knowledge in the visual-relational model (VRM) to equip the SGG model with the open-vocabulary capability.
 
 \paragraph{Standard finetuning strategy.}  It is intuitive to utilise some standard finetuning techniques~\cite{su2019vl,devlin2018bert}  to deploy a task-specific prediction head   for relation prediction and update   the pretrained VRM. 
 Following this setup, we could na\"ively design a finetuning strategy on the VRM. 
 
 Let $r_{so}$ denote the union region of      subject $r_s$ and object $r_o$, and $r_1,\ldots,r_m$ are the  object proposals overlapping with  $r_{so}$.  { We feed them into the pretrained image encoder as in Eq.(\ref{eq:trn})  to produce visual embeddings $\mathbf{h}$.}
 Then, we simply leverage  two classifiers to predict the predicate and object label by a cross-entropy loss:
 \begin{align}
 	\mathcal{L}_{f} &= \operatorname{CE}(\boldsymbol{W}\!_{r} \cdot \mathrm{h}_r)+ \sum_{h\in\{\mathrm{h}_s,\mathrm{h}_o\}}\operatorname{CE}(  \mathbf{W}\!_{c} \cdot  \mathrm{h}) 
 	\label{eq:ft}
 \end{align}
 where $\mathrm{h}_{r}$ is the embedding of the  union region of $r_s$ and $r_o$; $\mathrm{h}_{r}=\mathrm{LN}(\mathrm{h}_{s},\mathrm{h}_{so},\mathrm{h}_{o})$ where $\mathrm{LN}(.)$ is a linear project function; and $\boldsymbol{W}\!_{r}$ is the randomly initialized relation classifier. Different from relation classification, we utilize a zero-shot classification setting to predict object labels, that is, using the fixed embeddings of object categories  $\mathbf{W}\!_{c}$ by our pre-trained text encoder as the object classifier. In this way the model can predict unseen object categories; and $\operatorname{CE}$ is the cross-entropy function.  At finetuning,  all other parameters are updated, as in \cite{su2019vl}. 

 However, in practice, we find that the above  finetuning strategy does not gain satisfying performance for SGG, particularly in the open-vocabulary scenario. The main reason is updating all parameters could modified the preserved knowledge in the VRM and  damage the model's capability of generalization.
 
 \subsubsection{Prompt-based finetuning for Ov-SGG.}\label{sec:pro}
 Inspired by the success of prompt-based learning~\cite{trinh2018simple,petroni2019language,li2021prefix} on large-scale pre-trained language models such as GPT-3~\cite{brown2020language}, 
 we propose two prompt-based finetuning strategies, hard prompt and soft visual-relation prompt (SVRP), for Ov-SGG based on our pre-trained VRM. 
 A key advantage of our prompt-based tuning strategies is that they allow our pre-trained VRM to be optimized on task-specific data without updating its parameters. Doing so avoids altering the learned (open-vocabulary) knowledge during training, and thus supports predictions over unseen object labels at inference time.  
 
 The core of prompt-based learning is the design of \emph{templates} to convert the input sequence, $x_{in}$, into a textual prompt $\mathcal{X}_{pro}$ with unfilled  {cloze-style}   slots. The VRM then makes a prediction by filling the slot with the candidate of the maximum probability in the label space. 
 Without loss of generality, a prompt contains two key parts: a template $\mathcal{T}$ that can been manually designed~\cite{trinh2018simple} or learned from the training data~\cite{qin2021learning}; and a label mapping function $\mathcal{M}:\mathcal{Y}\rightarrow\mathcal{V}$ that maps a downstream task's labels $\mathcal{Y}$ to the vocabulary $\mathcal{V}$ of a pre-trained model.  {Note that in this work, $\mathcal{Y}=\mathcal{R}$ is the set of relation labels $\mathcal{R}$ as defined in Sec.~\ref{sec:pd}}, while $\mathcal{V}\!=\!\{w_i\}_{i=1}^{|\mathcal{V}|}$ is the set of words in the dense captions.
 
 \noindent \textbf{Hard prompt based finetuning.} Since relations in SGG are represented as SPO triples, we could readily formulate a relation prompt as follows:
 \begin{equation}
 	\mathcal{X}_{pro} = \mathcal{T}(x_{in})=\texttt{[CLS]} {x}_{\text {s }}\texttt{[MASK]} {x}_{\text {o}}\texttt{[SEP]}
 	\label{eq:templeate}
 \end{equation}
 where ${x}_{\text{s}}$ and ${x}_{\text{o}}$ denote the label of the subject and object, and $\texttt{[{MASK}]} $ is the slot for candidate predicate labels. Note that the labels of $x_{\text{s}}$ and $x_{\text{o}}$ are  produced in an zero-shot manner describing in the second term of Eq.~(\ref{eq:ft}). Then, we could predict the relation label by:
 \begin{equation}
 	{\hat{y}}= \underset{y \in \mathcal{Y}}{\operatorname{argmax}} ~   {P}\left(f_{\mathrm{fill}}\left( \mathcal{X}_{pro}, \mathcal{M}({y})\right) \mid \mathrm{h}_r ;  \boldsymbol{\theta}\right)
 	\label{eq:rel_pre}
 \end{equation}
 where $f_{\mathrm{fill}}(.)$ is a function that fills the slot $\texttt{[MASK]}$ in $\mathcal{X}_{pro}$ with a label word $\mathcal{M}({r}) \! \in \! \mathcal{V}$, $\mathrm{h}_r= \mathrm{LN}(\mathrm{h}_s,\mathrm{h}_{so},\mathrm{h}_o)$ is defined as in Eq.~(\ref{eq:ft}), and $\boldsymbol{\theta}$ are the parameters of the linear projection function $\mathrm{LN}$ (updated) and of VRM (frozen).
 For the prediction score $P$, we  also use Cosine similarity to calculate it: 
 \begin{equation}
 	P\left( {p} \mid \mathrm{h}_r, \mathcal{X}_{pro}\right)=\frac{\exp \left(\cos\left(\mathrm{h}_r , \mathrm{e}_\texttt{in}(p)\right)\right)}{\underset{q}{\sum}  \exp \left( \cos \left(\mathrm{h}_r , \mathrm{e}_\texttt{in}(q)\right)\right)}
 \end{equation}
 where $p=f_{\mathrm{fill}}(\mathcal{X}_{pro},\mathcal{M}(p))$ represents a filled prompt, $\mathrm{e}_\texttt{in}(p) = \operatorname{TexTrans}(p)$ denotes the textual embedding of $p$, and $q$ ranges over all filled prompts.
 
 \noindent \textbf{Soft visual-relation prompt (SVRP) based finetuning}.
 The above hard prompt employs a fixed template as shown in Eq.~(\ref{eq:templeate}), where only object label information is used. 
 In contrast, SVRP learns  a \emph{prefix}  visual-to-textual vector as the context to complement the hard prompt, that is,
 \begin{equation}
 	\mathcal{X}_{pro}^\prime  = {\texttt{{[\!CLS\!]}}\!\!   \texttt{[}{x}^{\prime}_s,\ldots, {x}^{\prime}_o\texttt{]}},  {{x}_{\text {s}} \texttt{[MASK]} {x}_{\text {o }}\! \texttt{[\!SEP\!]}}
 	\label{eq:prom_prim}
 \end{equation}
 where $\texttt{[}{x}^{\prime}_s,\ldots, {x}^{\prime}_o\texttt{]}$ denotes the prefix contextual vector. 
 
 To this end, we deploy a visual-to-textual decoder network $\mathbb{T}$  to decode the visual cues into the textual context, i.e.,
 $[{x}^{\prime}_s,\ldots, {x}^{\prime}_o]\! =\! \mathbb{T}(\mathbf{h})$, where $\mathbf{h} = \operatorname{[h_s,h_1,\ldots,h_m,h_o]}$ is produced by Eq.~(\ref{eq:trn}). Thus, we rewrite Eq.~(\ref{eq:prom_prim}) as: 
 $\mathcal{X}_{pro}^\prime =  \mathbb{T}(\mathcal{X}_{pro} \mid \mathbf{h})$. 
 Similar to Eq.~(\ref{eq:rel_pre}), the prediction can be formulated as: 
 \begin{equation}
 	{ \hat{y}}= \underset{y \in \mathcal{Y}}{\operatorname{argmax}} ~   {P}\left(f_{\mathrm{fill}}\left(  \mathcal{X}_{pro}^\prime, \mathcal{M}({y})\right); \boldsymbol{\theta}^\prime\right)
 \end{equation}
 where $\boldsymbol{\theta^\prime}$ are the trainable parameters of $\mathbb{T}$ and the fixed VRM.
 In this way, we view $ \mathcal{X}_{pro}^\prime$  as an input of a masked language model and  feed   it into the pre-trained $\operatorname{TexTrans}$ network to find the token that maximizes the probability of $\texttt{[MASK]}$. Thus, we could rewrite the prediction by:
 \begin{equation}
 	\begin{aligned} 
 		P\left(r \mid \mathcal{X}'_{pro} \right) &= P\left(\texttt{[MASK]}=\mathcal{M}(r) \mid \mathcal{X}_{pro}^\prime\right) \\ &=\frac{\exp \left(\cos \left(\mathbf{w}_{r} , \mathrm{e}_{\texttt {[MASK }]}\right)\right)}
 		{\underset{r^{\prime} \in \mathcal{R}}{\sum}
 			\!\!	\exp \left( \cos \left(\mathbf{w}_{r^{\prime}} , \mathrm{e}_{[\texttt{MASK}]}\right)\right)} \end{aligned}
 	\label{eq:xp}
 \end{equation}
 	where $\mathbf{w}_{r}$ is the embedding  of $r$ by   $\mathcal{E}_t$, and $\mathrm{e}_{\texttt{[MASK]}}$ is the  output of the token $\texttt{[MASK]}$ in $\mathcal{X}_{pro}^\prime$ by $\mathcal{E}_t$. During finetuning, we  inject the supervised examples \{($\boldsymbol{h}, \boldsymbol{y} $)\} to the model and utilise  cross-entropy loss to optimize Eq.~(\ref{eq:xp}).
 	
 \section{Experiments}\label{sec:exp}
 We evaluate our method on the task of SGG in both the new open-vocabulary and conventional closed-set settings on three benchmark datasets: Visual Genome (VG)~\cite{krishna2017visual}, GQA~\cite{hudson2019gqa} and Open-Image~\cite{kuznetsova2020open}. Implementation details can be found in the supplementary materials.

 \subsection{Datasets}
 \noindent \textbf{Visual Genome} (VG) is the mainstream benchmark dataset for SGG. Following previous works~\cite{zellers2018neural,tang2020unbiased,xu2017scene}, we use the pre-processed VG with $150$ object classes and $50$ predicates~\cite{xu2017scene}. VG consists of 108k images, of which $57,723$ images are used for training and $26,443$ for testing. Additionally, $5,000$ images make up the validation set.
 
 \noindent \textbf{GQA} is a more challenging dataset derived from VG images. It contains $1,704$ object categories and $311$ predicate words. Following  \cite{knyazev2020graph},  we have  $66,078$ training images, $4,903$ validation images, and $10,055$ test images. 
 
 \noindent \textbf{Open-Image}(v$_6$) consists of $301$ object
 categories and $31$ predicate categories. Following the split of \cite{han2021image}, the training set has 
 $126,368$ images while the validation and test sets contain $1,813$ and $5,322$ images respectively. 

 \subsection{Evaluation Settings}
 We evaluate model performance in  Ov-SGG as well as two other related settings: closed-set SGG and {zero-shot object SGG}. 
 Before training, we fist randomly split all the object classes into two groups, \emph{base classes} and \emph{target classes}, on each experimental dataset, with $70$\% objects for the base group whilst the remaining $30$\% for the target. 
 
 \noindent \textbf{Closed-set SGG (Cs-SGG)} is the conventional standard SGG evaluation protocol in which a model predicts the relation between base objects only, i.e., $\mathcal{O}^b\times\mathcal{O}^b$. 
 We report the results on  the two subtasks: Predicate Classification (\textsc{PredCls}) and  Scene Graph Classification (\textsc{SGCls}).  
 
 \noindent \textbf{Open-vocabulary SGG (Ov-SGG)} aims to evaluate the model's ability to recognize the relationship between open-vocabulary objects, i.e., $\mathcal{O}\times\mathcal{O}$. 
 {We discard the third subtask \textsc{SGDet} in this setting, as the   object detection network \cite{zhang2021vinvl} can not handle open-vocabulary    object detection.} 
 
 \noindent \textbf{Zero-shot Object SGG (ZsO-SGG)} is different from the previous Zero-shot SGG \cite{tang2020unbiased,suhail2021energy}, which simply aims to evaluate a model's capability of predicting unseen object pairs, i.e., the combinations of subject and object classes has not appeared in the training set. In contrast, we set the zero-shot configuration on the object level, that is, to predict the predicate between two object classes completely unseen during training, i.e., $\mathcal{O}^t\times\mathcal{O}^t$. Thus, this setting can be regarded as a special case of Ov-SGG.

 \noindent \textbf{Metrics}. For VG and GQA, we mainly report Recall@K (R@K). As the bias of R@K has been widely acknowledged~\cite{suhail2021energy,tang2020unbiased}, we further report the results on mR@K. For Open-Image, following the settings of \cite{han2021image}, we report three metrics: Recall@50, weighed mean Average Precision (wmAP) and mean Average Precision (mAP) of triples. 
 
 \noindent \textbf{Baselines.} We compare with recent and strong SGG methods: VTransE~\cite{zhang2017visual}, IMP~\cite{xu2017scene}, Motifs \cite{zellers2018neural}, VCtree~\cite{tang2019learning}, TDE~\cite{tang2020unbiased}, GCA~\cite{Knyazev_2021_ICCV},  and EBM~\cite{suhail2021energy}. Since TDE and EBM are model-free, for a fair comparison, we choose VCTree as their baseline model to apply their techniques.  
 
 

 \begin{table*}[htb]
 	
 	\centering
 	\resizebox{0.97\linewidth}{!}
 	{ 
 		\begin{tabular}{llcccccccccccc}
 			\toprule
 			
 			\multirow{12}{*}{\rotatebox{270} {\textbf{VG}}}	   & \multirow{3}{*}{Models} & \multicolumn{4}{c}{Cs-SGG(70\%)}                                                                      &  \multicolumn{4}{c}{Ov-SGG(70\%+30\%)}                                                          & \multicolumn{4}{c}{ZsO-SGG(30\%)}                                     \\
 			&                           & \multicolumn{2}{c}{\textsc{PredCls}}                      & \multicolumn{2}{c}{\textsc{SGCls}}                        & \multicolumn{2}{c}{\textsc{PredCls}}                    & \multicolumn{2}{c}{\textsc{SGCls}}                        & \multicolumn{2}{c}{\textsc{PredCls}} & \multicolumn{2}{c}{\textsc{SGCls}} \\
 			&                  & R@50                   &  {R@100}   &  {R@50} &  {R@100} & R@50                   &  {R@100} &  {R@50} &  {R@100} & R@50           & R@100          & R@50          & R@100         \\ \cmidrule{2-14}
 			& IMP~\cite{xu2017scene}                     & 46.93	&  48.12   & 28.20  & 28.91   & 40.02  & 43.40  & 23.86  & 25.71  & 37.01 & 39.46   &-  &  -    \\
 			& Motifs \cite{zellers2018neural}            & 49.41	&  50.71   & 29.65  & 30.43   & 41.14  & 44.70  & 24.02  & 27.12  & 39.53 & 41.14  &-  &  -  \\
 			& VCtree~\cite{tang2019learning}             & 50.13    &  52.50   & 30.09  & 31.02   & 42.56  & 45.84  & 25.24  & 28.47  & 41.27 & 42.52  &-  &  -  \\
 			& TDE~\cite{tang2020unbiased}                & 45.21    &  46.03   & 27.14  & 27.64   & 38.29  & 40.38  & 22.56  & 24.22  & 34.15 & 36.37  &-  &  -   \\
 			& GCA~\cite{Knyazev_2021_ICCV}               & 51.15    &  53.38   & 29.89  & 32.10   & 43.48  & 46.26  & 25.71  & 27.40  & 42.56 & 43.18  &-  &  -  \\ 
 			& EBM~\cite{suhail2021energy}                & 52.81    &  54.91   & 31.64  & 32.95   & 44.09  & 46.95  & 26.03  & 28.03  & 43.27 & 44.03  &-  &  -   \\ 
 			\cmidrule{2-14}
 			& \textbf{SVRP}                              & \textbf{54.39}    &  \textbf{56.42}   & \textbf{32.75}  & \textbf{33.87}   & \textbf{47.62}  & \textbf{49.94}  & \textbf{28.40}  & \textbf{30.13}  & \textbf{45.75} & \textbf{48.39}  & \textbf{9.30}  &  \textbf{11.32}   \\
 			\multirow{-1.3}{*}{\rotatebox{270}{Ablations}}	 	
 			& \quad FT-p		                         & 48.13    &  50.05   & 28.82  & 29.12   & 42.13  & 45.10  & 24.47  & 26.64  & 41.02  & 44.10  &-  &  -   \\
 			& \quad FT                             & 51.42    &  53.51   & 29.13  & 30.46   & 44.73  & 46.49  & 26.24  & 28.02  &  42.86  & 46.25  & 5.16  &  7.42   \\
 			& \quad HardPro                              & 46.83    &  54.12   & 30.93  & 32.41   & 46.34  & 48.02  & 26.87  & 29.41  & 43.42 & 47.05  & 6.61  &  8.13   \\
 			& \quad SVRP-d		                         & 47.05    &  54.27   & 31.05  & 32.20   & 46.96  & 48.29  & 27.68  & 29.89  & 44.31 & 47.53  & 7.84  &  10.25   \\

 			\midrule
 			\specialrule{0em}{1pt}{1pt}
 			\midrule

 			\multirow{5}{*}{\rotatebox{270} {\textbf{GQA}}}
 			& IMP~\cite{xu2017scene}                     &  50.73  &  54.44   & 17.28    & 20.16   &  30.60 & 34.30  & 9.61  & 11.86  &  24.34 &  28.85  &-  &  -        \\
 			& Motifs \cite{zellers2018neural}            &  51.74  &  56.10   & 18.93    & 21.02   &  32.24 & 36.04  & 12.37  & 13.40  & 28.24 & 31.34  &-  &  -         \\	
 			& VCtree~\cite{tang2019learning}             &  51.02  &  55.78   & 19.31    & 21.76   &  32.06 & 36.86  & 12.86  & 13.73  & 30.83 & 33.17  &-  &  -         \\
 			& TDE~\cite{tang2020unbiased}                &  48.30  &  52.69   & 16.45    & 18.50   &  29.15 & 33.37  & 10.36  & 11.42  & 24.16 & 26.83  &-  &  -         \\
 			& GCA~\cite{Knyazev_2021_ICCV}               &  53.83  &  57.85   & 20.73    & 22.84   &  33.47 & 37.11  & 13.13  & 14.57  & 32.73 &  34.24   & -  &  -        \\
 			& EBM~\cite{suhail2021energy}                &  53.27  &  58.16   & 20.01    & 22.52   &  33.03 & 36.82  & 13.01  & 14.34  & 32.10 & 34.65  &-  &  -        \\ 
 			\cmidrule{2-14}
 			& \textbf{SVRP}                   			 &  \textbf{55.85}  &  \textbf{61.21}   & \textbf{22.54}    & \textbf{24.38}   &  \textbf{35.26} & \textbf{39.03}  & 
 			\textbf{15.16}  & \textbf{16.42}  & \textbf{34.70} & \textbf{36.79}  & \textbf{1.72} & \textbf{2.35}         \\

 			\multirow{-1.3}{*}{\rotatebox{270}{Ablations}}
 			& \quad FT-p		     					 &  50.22  &  54.93   & 17.38    & 19.11   &  31.37 & 35.02  & 11.22  & 12.46  & 27.58 & 30.13  &-  &  -         \\
 			& \quad FT                   				 &  52.14  &  56.17   & 19.49    & 21.63   &  33.08 & 36.77  & 13.35  & 14.38  & 29.51 & 32.42  & ~.19 & ~.80         \\
 			& \quad HardPro                  		     &  54.81  &  58.92   & 20.10    & 22.03   &  34.19 & 37.05  & 14.18  & 14.90  & 32.64 & 34.01  & 1.04 & 1.43          \\
 			& \quad SVRP-d                  			 &  55.13  &  59.86   & 20.52    & 22.75   &  34.62 & 37.98  & 14.75  & 15.42  & 33.53 & 35.24  & 1.12 & 1.72          \\	
 			
 			\bottomrule
 		\end{tabular}
 	}
 	\caption{The comparison Recall@K results of  Visual Genome and GQA datasets with other state-of-the-arts SGG models on the tasks of  closed SGG, ov-SGG and ZsO-SGG, where we use 70\% object for training and the remaining 30\% as the unseen objects for testing. All compared methods use the same backbone of ResNet-50 as in \cite{han2021image}. `-' indicates models incapable of obtaining the result. Note that  following  \cite{knyazev2020graph}, we report the results of GQA under the setting of graph unconstraint. 
 	}
 	\label{tab:1}
 \end{table*}
 
 \subsection{Results and Analysis}
 Tab. \ref{tab:1} presents the comparison results on VG and GQA with other state-of-the-arts models on the three tasks: Cs-SGG, Ov-SGG and ZsO-SGG. It is worth noting that for the task of ZsO-SGG, the conventional SGG models are incapable of predicting object labels of unseen classes and thus we do not report their results. For  GQA, we follow the setting of \cite{knyazev2020graph} and     compute recalls without the graph constraint. 
 
 \noindent \textbf{Cs-SGG.} In this conventional setting, we could observe that our SVRP on both VG and GQA consistently surpasses all the compared baselines, even the recent, SOTA models GCA~\cite{Knyazev_2021_ICCV} and EBM~\cite{suhail2021energy}. For instance, SVPR on VG  gains averagely about {$1.55$ and  $1.02$ points}   of improvements on the task of \textsc{PredCls} and \textsc{SGCls} when compared to EBM. Compared to the other models, e.g., IMP and Motifs, SVPR exceed them by even larger margins, e.g., about  $6.32$   points better than IMP on average.   
 
 \begin{table}[!ht]
 	\centering
 		\begin{tabular}{lcccccc}
 			\toprule
 			\multirow{2}{*}{Models} & \multicolumn{3}{c}{Cs-SGG(70\%)} & \multicolumn{3}{c}{Ov-SGG(70\%+30\%)}                                         \\
 			& R@50    &  mAP   &     wmAP   & \multicolumn{1}{c}{R@50} &  \multicolumn{1}{c}{mAP}  &  \multicolumn{1}{c}{wmAP}\\  \midrule
 			IMP~\cite{xu2017scene}             &  71.54   &   34.67   & 30.01  &   51.38 &  24.15   &  21.62     \\
 			Motifs \cite{zellers2018neural}    &  72.05   &   32.70   & 29.16  &   50.47 &  23.03   &  20.51     \\
 			VCtree~\cite{tang2019learning}     &  72.41   &   33.08   & 30.15  &   52.12 &  24.29   &  22.14      \\
 			TDE~\cite{tang2020unbiased}        &  70.52   &   30.10   & 31.68  &   49.50 &  21.38   &  18.33    \\
 			EBM~\cite{suhail2021energy}        &  71.26   &   34.46   & 30.25   &  52.03 &  24.59   &  23.31    \\
 			\midrule
 			\textbf{SVRP}                      &  \textbf{72.84}   &   \textbf{35.32}   & \textbf{33.11}   &   \textbf{54.45} &  \textbf{27.97}   &  \textbf{25.36}   \\
 			\bottomrule
 		\end{tabular} 
 	\caption{Results on Open-Image(v6) in the closed-set and open-vocabulary settings. All methods used the X152FPN network~\cite{han2021image} pretrained on Open Images.  }
 	\label{tab:oi}
 \end{table}
 \noindent \textbf{Ov-SGG.} The middle block shows the results of the new, challenging Ov-SGG setting, in which all models suffer considerable performance drops  when compared to Cs-SGG. However, our SVRP technique still obtains the best results, on average  {$3.91$ and $2.71$ points}   higher than GCA in terms of \textsc{PredCls} and \textsc{SGCls}, respectively. The conventional models such as Motifs and VCTree, which were designed for Cs-SGG, struggle on this task. We posit the main reason to be their reliance on manually-designed external knowledge, e.g., word embeddings from \textsc{Glove}~\cite{pennington2014glove}. Although beneficial for Cs-SGG, this knowledge damages the model's generalizability for open-vocabulary objects, as they do not see the embeddings of the target object during training, and are prone to overfitting on the base object classes. In contrast, through pre-training on massive dense-caption corpus, our pre-trained VRM directly learns to align the visual and relation knowledge, which can avoid the overfitting problem on the base classes. Additional, our prompt-based mechanisms allow us to finetune VRM without modifying its parameters, which makes it possible to extend the generalizability of VRM to Ov-SGG.

 \noindent \textbf{ZsO-SGG.} The rightmost block shows the results of ZsO-SGG. Again, our SVRP is superior to all baseline models in \textsc{PredCls} on the two datasets. In particular, SVRP on average exceeds the strong EBM and GCA by about 3.42 and 4.20, respectively. Moreover, our method is the only one that is capable of handling SGCls in this setting. 
 \begin{table}[!ht]
 	\centering
 	{ 
 		\begin{tabular}{lcccccc}
 			\toprule  
 			\multirow{2.9}{*}{{Models}} & \multicolumn{2}{c}{{{PredCls}}}   & \multicolumn{2}{c}{{{SGCls}}}     & \multicolumn{2}{c}{{{SGDet}}}    \\  

 			& {{R@K}} & {{mR@K}} & {{R@K}} & {{mR@K}} & {{R@K}} & {{mR@K}} \\ \midrule %
 			IMP~\cite{xu2017scene}            & 54.1/57.9 &  9.3/10.2  &  30.3/32.2          &  5.8/6.2   &  19.7/24.6    & 1.3/7.6           \\ 
 			Motifs \cite{zellers2018neural}    & 60.1/61.5 &  13.8/15.1 &  32.1/34.5          &  7.3/8.0   &  26.2/28.3    &  5.2/6.3                 \\ 
 			VCtree~\cite{tang2019learning}     & 59.6/60.4 &  16.7/18.4 &  33.0/35.7          &  9.1/10.3  &  25.9/30.0    &  8.2/10.1                \\ 
 			TDE~\cite{tang2020unbiased}        & 50.2/55.8 &  20.3/24.2 &  27.2/30.4          &  10.4/12.5 &  22.6/25.9    &  8.6/10.5                \\ 
 			GCA~\cite{Knyazev_2021_ICCV}       & 58.9/60.9 &  22.2/23.3 &  29.1/33.3          &  11.4/13.1 &  -/-     &     -/-              \\ 
 			
 			EBM~\cite{suhail2021energy}        & 59.8/61.7 &  24.1/\textbf{26.0} &  32.0/34.6    &  13.1/14.9 &  26.8/33.7    & 9.2/11.6                \\%
 			
 			\cmidrule{1-7} 
 			
 			\textbf{SVRP}                       & \textbf{60.2}/\textbf{62.3} &  \textbf{24.3}/25.3 &  \textbf{33.9}/\textbf{35.2}          & \textbf{12.5}/\textbf{15.3} &  \textbf{31.8}/\textbf{35.8 }   & \textbf{10.5}/\textbf{12.8}                \\    
 			
 			\quad {FT-p}							& 56.0/60.5 &  20.1/23.6 &  31.7/32.1          &  10.1/11.4 &  26.7/32.5    & 7.3/9.4                  \\ 
 			\quad FT       							& 58.8/62.2 &  22.7/24.5 &  32.4/34.8          &  11.3/13.5 &  27.0/33.2    & 9.0/10.5                \\  
 			\quad HardPro       					& 60.5/63.1 &  23.2/25.0 &  32.1/34.5          &  13.2/14.9 &  30.2/34.1    & 9.8/12.0                \\ 
 			\quad SVRP-d     						& 60.2/62.7 &  23.8/25.3 &  33.5/35.0          &  14.0/15.2 &  31.3/34.8    & 10.2/12.7                \\ 
 			
 			\bottomrule
 		\end{tabular}
 		`	}
 	\caption{The results of fully closed scene graph generation on VG and  $\mathrm{K}$ is set to ${50/100}$. All compared methods use the object detection network from \cite{han2021image} with ResNet-50 as the backbone. }
 	\label{tab:fcsgg}
 \end{table}
 Tab. \ref{tab:oi} shows the results on Open-Image. Interestingly, we could find that the performance of ZsO-SGG is significantly higher than the conventional Zs-SGG~\cite{tang2020unbiased}, possibly because the majority of predicates in the ZsO-SGG are frequent  relationships and this is in favor of better results with the biased metric R@K.  
 
 \noindent \textbf{Fully-closed scene graph generation.}
 We further evaluate our technique for fully closed scene graph generation, as shown in Table \ref{tab:fcsgg}. Specifically, we use all object classes in VG to train following the conventional setting~\cite{tang2019learning}. We discard all the statistical bias prior information for all compared methods. 
 From the results, we could observe that our SVRP surpasses all compared baselines, except for mR@100 on the task of \textsc{PredCls}, our SVRP has $0.7$ points lower than EBM.

 \noindent \textbf{Ablations.} We evaluate the effectiveness of our model components in several variants: (1) $\operatorname{FT-p}$ uses the pretrained model from \cite{zhang2021vinvl} instead of our VRM, i.e., without pre-training; (2) $\operatorname{FT}$ uses the standard finetuning strategy described in Sec.\ref{sec:finetune}; (3) $\operatorname{HardPro}$ uses hard prompt finetuning described in Sec.~\ref{sec:pro}; and (4) $\operatorname{SVRP-d}$ removes the decoder network $\mathbb{T}$ for SVRP. The ablation results on the VG and GQA datasets are shown in Tab.~\ref{tab:1} and \ref{tab:fcsgg}, respectively. With the pre-trained VRM removed (FT-p $vs.$ FT), we could find that our VRM can bring 2$\sim$3 points improvements, which suggests that simply using the visual-language model \cite{zhang2021vinvl} for SGG does not bring much benefit, possibly because the visual-language model trained on the global image-caption pairs focuses on the images' global semantics, but ignores the regional semantics. However, those regional cues play an important role in SGG. 
 Furthermore, our two prompt-based finetuning techniques present clear superiority to the standard finetuning strategy, especially in the open-vocabulary scenarios, since the prompt-based strategies directly leverage knowledge preserved in the pre-trained model, which equips the downstream SGG model with the zero-shot capability, whilst the standard finetuning strategy updates the pre-trained model and therefore creates interference between tasks. 
 With respect to the decoder network ($\operatorname{SVRP-d}$), the results confirm  solely feeding  plain region embeddings to the language model does not generate good prefix contexts for prompts. 
 \begin{table}[ht]
 	\centering
 		\begin{tabular}{l|cccc|cccc}
 			\toprule  
 			\multirow{2.9}{*}{{Models}} & \multicolumn{4}{c|}{{{PredCls}}}   & \multicolumn{4}{c}{{{SGCls}}}     \\ 
 			& {{R@50}}& R@100 & {{mR@50}} &mR@100 & {{R@50}}& R@100& {{mR@50}} & mR@100 \\ \midrule %

 			\textbf{SVRP}  & \textbf{33.5}&\textbf{35.9} &  \textbf{8.3}&\textbf{10.8} &  \textbf{19.1}&\textbf{21.5}          &  {3.2}&\textbf{4.5}             \\    
 			
 			\quad {FT-p}							& 29.4&31.3 &  5.8& 6.7 &  16.9& 18.0          &  2.3&2.7      \\ 
 			
 			\quad HardPro       					& 30.0  &32.6 &  6.9&8.1 &  17.2&18.6           &   3.0& 4.1         \\ 
 			\quad SVRP-d   	& 31.3&33.2 &  7.6&9.0 &   18.3& 19.7          &  3.2 & 4.4               \\ 
 			
 			\bottomrule
 		\end{tabular}
 	\caption{The results of  gOv-SGG on VG where $\mathrm{K}$ is set to ${50/100}$.  }
 	\label{tab:gov-sgg}
 \end{table}
 \begin{figure*}[!h]
 	\centering
 	\includegraphics[width=0.95\linewidth]{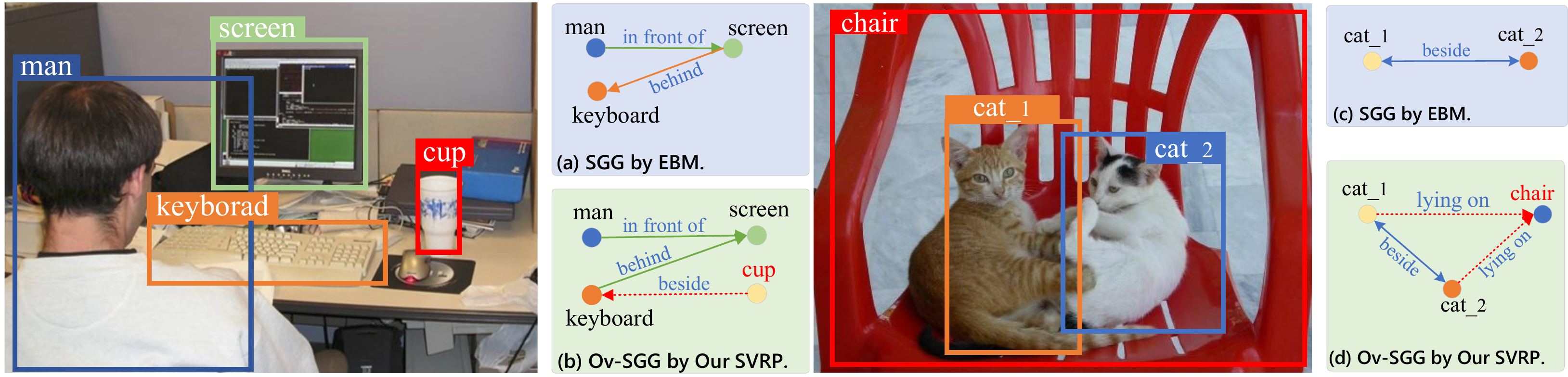}
 	\caption{Qualitative results of our SVRP and EBM~\cite{suhail2021energy} on the VG test set for Ov-SGG. EBM is unable to detect relations on the unseen `cup' and 'chair' classes while our method can.}  \label{fig:qa}
 \end{figure*}
 
 \noindent \textbf{gOv-SGG.} Additionally, we also provide a baseline for the challenging task of gOv-SGG, in which the model needs to make predictions on novel relation predicates during inference. In this setting, the model sees 70\% objects and 70\% predicates in the training stage. Specifically, the finetuning predicate set $\mathcal{R}$ in Eq.~(\ref{eq:xp})  only account for 70\% of all predicates. The model then tries to predict the remaining 30\% predicates words on the whole of objects. Since none of the baseline methods can handle this task, we only report the results of our models and its ablated variants, as shown in Table~\ref{tab:gov-sgg}. We could obviously see that our SVPR still achieves the best results over all metrics.
 
 

 \noindent \textbf{Qualitative analysis.}
 On the task of Ov-SGG, we visualize the scene graphs produced by our technique as well as by the representative closed SGG model EMB~\cite{suhail2021energy}, as shown in Fig. \ref{fig:qa}. For the left image, EBM cannot detect any relation regrading the unseen target object ``cup'', while our SVRP can predict the ``beside'' relationship between the keyboard and cup. Similarly for the right image, EBM is unable to make predictions about the unseen object class ``chair'' while our method can. 

 \section{Conclusion}
 We propose the new practical and challenging  open-vocabulary scene graph generation (Ov-SGG) setting and design a two-step method that firstly pre-trains a visual-relation model on large-scale region-caption data. We develop two prompt-based strategies to finetune the visual-relation model for the downstream Ov-SGG task without modifying the pret-rained model parameters. Our extensive experiments on three benchmark datasets show that our method significantly outperforms recent, strong SGG methods on the setting of Ov-SGG, gOv-SGG and the closed SGG.  In the future, we hope to integrate   open-vocabulary object detection into Ov-SGG.
 
 \noindent\textbf{Acknowledgments. } This work was partially funded by DARPA CCU program (HR001121S0024).
 
\bibliographystyle{splncs04}
\bibliography{egbib}

\section{Supplementary} \label{sec:model}
\subsection{ More Model Details}
We use the cosine contrastive loss to optimize region-caption pair following \cite{radford2021learning}. More specifically, we can write the match loss as the below:
\begin{equation}
	\begin{aligned} 
		\mathcal{L}_{c}\! =\!\!&\sum_{i \in \operatorname{U(r_t,r_b)}} \!\sum_{j \in \operatorname{U(r_t,r_b)}} \!\!\!\!\! \mathbbm{1}_{y_{j}^{i}=1} \log \!\left(\!\frac{1}{1+e^{-\tau \cos(\boldsymbol{e}_{i}, \mathbf{h}_{j})}}\!\right) \\ + \!\!\! &\sum_{i \in \operatorname{U(r_t,r_b)}} \!\sum_{j \in \operatorname{U(r_t,r_b)}}\!\!\!\!\!\mathbbm{1}_{y_{t}^{i}=0} \log \!\left(\!\frac{1}{1+e^{\tau \cos( \boldsymbol{e}_{i}, \mathbf{h}_{j})}}\!\right)   
	\end{aligned}
	\label{eq:contras}
\end{equation}
where $\boldsymbol{e}_i$ and $\mathbf{h}_j$ are a caption embedding and visual region embedding, and produced by the text encoder $\mathcal{E}_t$ and image encoder $\mathcal{E}_i$, respectively; $r_t$ and $r_b$ are selected anchor regions, i.e., the top left region and bottom right region respectively;  $\mathbbm{1}$ is an indicator function  and if a visual region and a textual caption are matched, its value is $1$, otherwise $0$; $\operatorname{U(r_t,r_b)}$ denotes  all regions overlapped with the union region of $r_t$ and $r_b$, $\cos(.)$ is the cosine function and $\tau$ is a temperature parameter jointedly learned.

For the masked region loss $\mathcal{L}_\mathrm{MRL}$, we also use the contrastive loss to optimize it, but the difference is that a visual region $i \in \operatorname{U(r_t,r_b)}$  may be replaced by a masked region with  $15$\% possibilities. Formally, we could formulate it as:
\begin{equation}
	\begin{aligned} 
		\mathcal{L}_\mathrm{MRL}\! =\!\!&  \!\sum_{j \in \operatorname{U(r_r,r_b)}} \!\!\!\!\! \mathbbm{1}_{y_{j}^{i}=1} \log \!\left(\!\frac{1}{1+e^{-\tau \cos(\boldsymbol{e}_{j}, \mathbf{h}_{m})}}\!\right) \\ + \!\!\! &  \!\sum_{j \in \operatorname{U(r_t,r_b)}}\!\!\!\!\!\mathbbm{1}_{y_{t}^{i}=0} \log \!\left(\!\frac{1}{1+e^{\tau \cos( \boldsymbol{e}_{j}, \mathbf{h}_{m})}}\!\right)   
	\end{aligned}
	\label{eq:mrl}
\end{equation}
where $\mathbf{h}_m$ is the masked visual region. In our implementation, we use a learnable embedding vector to replace its original visual feature. 

Since the  localisation  of visual regions  in an image can provide essential cues for relation recognition, we also add the spatial information to the pretraining of VRM. Specifically, for each region $r_j$, we extract its four coordinates $\mathbf{s}_j = [x^j_t, y^j_t, x^j_b, y_b^j ]$, where $\__t$ and $\__b$ denote the top-left and bottom-right coordinate respectively. Then, we use a project module to transform it into a $128$-dimensional informative spatial feature vector   and concatenate it with its corresponding  visual feature.   During finetuning, we treat $\mathcal{M}(\mathcal{R})$ as $\mathcal{R}$, that is, $r=\mathcal{M}(r)$. Thus, we do not conduct the relation label search processing for our prompt.  Additionally, for the task of \textsc{SGCls}, we leverage a zero-shot object classification strategy to predict  object labels \cite{rahman2018zero}.

 \subsection{Implementation Details}\label{sec:imp}

During pretraining  the VRM,  we fix the convolutional backbone of Faster-RCNN with the architecture of  ResNet-50   but only train the region feature extractor module. And we discard the region proposal network (RPN) in the pretraining, since we do not consider to locate objects in an image.  
For the more challenging open-image dataset, we use X152FPN   as the backbone network.
As for the image encoder, we randomly sample $10$ union regions for each image, and we set an IoU threshold $0.4$ to select those overlapped regions with the union region as the context. Furthermore, we fix the CNN backbone and only train the Transformer network, which consists of 8 self-attention blocks with 8 attention heads. The vocabulary size of the text encoder is set to $49,152$ and the dimension of word embeddings is set to $768$.   The final embedding dimension for both encoders is $512$. We use SGD with an initial learn learning rate 0.01 adjusted by a WarmupMultiStepLR scheduler to optimize the loss in Eq.~(3) with the batch size set to $3$, which results in approx.\ $100$ region-description pairs for each mini batch. The split of the object classes are shown in Table \ref{tab:split}.   
\begin{table}[]
	\centering
	\caption{The split of seen objects and unseen object during training. }
	\label{tab:split}
		\resizebox{1 \columnwidth}{!}{ 
	\begin{tabular}{l|l}
		\toprule
		& \multicolumn{1}{c}{Names of Object Classes}                                                                                                                                                                                                                                                                                                                                                                                                                                                                                                                                                                                                                                                                                                                                                                                                                                                                                                                                                                                                                \\ \cmidrule{1-2}
		Seen objects   & \begin{tabular}[c]{@{}l@{}}'tile', 'drawer', 'men', 'railing', 'stand', 'towel', 'sneaker', 'vegetable', \\'screen',  'vehicle', 'animal', 'kite', 'cabinet', 'sink', 'wire', 'fruit', 'curtain', \\'lamp', 'flag', 'pot', 'sock', 'boot', 'guy', 'kid', 'finger', 'basket', 'wave', 'lady', \\ 'orange', 'number', 'toilet', 'post', 'room', 'paper', 'mountain', 'paw', \\'banana', 'rock', 'cup', 'hill', 'house', 'airplane', 'plant', 'skier', 'fork', \\'box', 'seat', 'engine', 'mouth', 'letter', 'windshield', 'desk', 'board', 'counter',\\ 'branch', 'coat', 'logo', 'book', 'roof',  'tie', 'tower', 'glove', 'sheep', 'neck', \\'shelf', 'bottle', 'cap', 'vase', 'racket', 'ski',  'phone', 'handle', 'boat', 'tire', \\'flower', 'child', 'bowl', 'pillow', 'player', 'trunk',  'bag', 'wing', 'light', 'laptop',\\ 'pizza', 'cow', 'truck', 'jean', 'eye', 'arm', 'leaf',   'bird', 'surfboard', 'umbrella', \\'food', 'people', 'nose', 'beach', 'sidewalk',   'helmet', 'face', 'skateboard', \\ 'motorcycle', 'clock', 'bear'\end{tabular} \\ \midrule
		Unseen objects & \begin{tabular}[c]{@{}l@{}}'bed', 'track', 'shoe', 'wheel', 'chair', 'bench', 'short', 'bike', 'fence', 'door', \\ 'zebra', 'jacket', 'car', 'pant', 'street', 'horse', 'plane', 'snow', 'tail', 'pole', \\ 'hat', 'glass', 'girl', 'dog', 'elephant', 'giraffe', 'bus', 'plate', 'ear', 'cat', \\ 'sign', 'hand', 'hair', 'boy', 'train', 'leg', 'head', 'tree', 'table','building', \\ 'window', 'shirt', 'person', 'woman', 'man'\end{tabular}  \\
		\bottomrule
		\end{tabular}
}
\end{table}

\end{document}